\documentclass{article}

\usepackage{arxiv}

\usepackage[utf8]{inputenc} 
\usepackage[T1]{fontenc}    
\usepackage{hyperref}       
\usepackage{url}            
\usepackage{booktabs}       
\usepackage{amsfonts}       
\usepackage{nicefrac}       
\usepackage{microtype}      
\usepackage{lipsum}
\usepackage{multirow}
\usepackage{subfigure}
\usepackage{graphicx}
\usepackage{footnote}
\makesavenoteenv{tabular}
\makesavenoteenv{table}

\title{NEZHA: Neural Contextualized Representation for Chinese Language Understanding}

\author{
 Junqiu Wei, Xiaozhe Ren, Xiaoguang Li, Wenyong Huang, Yi Liao, \\
 \textbf{Yasheng Wang, Jiashu Lin$^*$, Xin Jiang, Xiao Chen, Qun Liu} \\
 Noah's Ark Lab, $^*$HiSilicon, Huawei Technologies\\
   \texttt{\{wei.junqiu1, renxiaozhe, lixiaoguang11, wenyong.huang, liao.yi, }\\ \texttt{wangyasheng, linjiashu, jiang.xin, chen.xiao2, qun.liu\}@huawei.com} \\
}

\begin{document}
\maketitle

\begin{abstract}
The pre-trained language models have achieved great successes in various natural language understanding (NLU) tasks due to its capacity to capture the deep contextualized information in text by pre-training on large-scale corpora. In this technical report, we present our practice of pre-training language models named NEZHA (\underline{NE}ural contextuali\underline{Z}ed representation for C\underline{H}inese l\underline{A}nguage understanding) on Chinese corpora and finetuning for the Chinese NLU tasks. The current version of NEZHA is based on BERT~\cite{devlin2019BERT} with a collection of proven improvements, which include \emph{Functional Relative Positional Encoding} as an effective positional encoding scheme, \emph{Whole Word Masking} strategy, \emph{Mixed Precision Training} and the \emph{LAMB Optimizer} in training the models. The experimental results show that NEZHA achieves the state-of-the-art performances when finetuned on several representative Chinese tasks, including named entity recognition (People's Daily NER), sentence matching (LCQMC), Chinese sentiment classification (ChnSenti) and natural language inference (XNLI). 
\end{abstract}


\section{Introduction}

Pre-trained language models such as ELMo~\cite{peters2018deep}, BERT~\cite{devlin2019BERT}, ERNIE-Baidu~\cite{sun2019ernie,sun2019ernie2}, ERNIE-Tsinghua~\cite{zhang2019ernie},
XLNet~\cite{yang2019xlnet}, RoBERTa~\cite{Liu2019RoBERTaAR} and MegatronLM\footnote{\url{https://nv-adlr.github.io/MegatronLM}} have demonstrated remarkable successes in modeling contextualized word representations by utilizing the massive amount of training text. As a fundamental technique in natural language processing (NLP), the language models pre-trained on text could be easily transferred to learn downstream NLP tasks with finetuning, which achieve the state-of-the-art performances on many tasks including sentiment analysis, machine reading comprehension, sentence matching, named entity recognition and natural language inference.

The existing pre-trained language models are mostly learned from English corpora (e.g., BooksCorpus and English Wikipedia). There are several attempts to train the models specifically for the Chinese language, including Google's BERT~\cite{devlin2019BERT} for Chinese, ERNIE-Baidu~\cite{sun2019ernie,sun2019ernie2} and BERT-WWM~\cite{cui2019pre}. All of the models are based on Transformer~\cite{vaswani2017attention} and trained on two unsupervised tasks: \emph{Masked Language Model (MLM)} and \emph{Next Sentence Prediction (NSP)}. In the MLM task, the model learns to recover the masked words in the training sentences. In the NSP task, it tries to predict whether one sentence is the next sentence of the other. One of the main differences among the Chinese models lies their word masking strategy in the MLM task. Google's BERT masks each Chinese character or WordPiece token~\cite{wu2016google} independently. ERNIE-Baidu further makes the MLM task more challenging by masking the entities or phrases in a sentence as a whole, where each entity or phrase may contain multiple characters or tokens. BERT-WWM takes a similar strategy called \emph{Whole Word Masking (WWM)}, which enforces that all the tokens belonging to a Chinese word should be masked together. Besides, in the most recently published ERNIE-Baidu 2.0~\cite{sun2019ernie2}, additional pre-training tasks such as \emph{Token-Document Relation Prediction} and \emph{Sentence Reordering}, are also incorporated.

In this technical report, we present our practice of pre-training language models NEZHA (\underline{NE}ural contextuali\underline{Z}ed representation for C\underline{H}inese l\underline{A}nguage understanding), which is currently based on BERT and trained on Chinese text. Specifically, we employ a technique called \emph{Functional Relative Positional Encoding} in our model. In the vanilla Transformer as well as the BERT model, the positional encoding of each word in the sequence is a vector with its absolute position information encoded. The positional encodings are added to the word embeddings as the inputs to the Transformer. There are two typical ways to determine the positional encodings. One is the \emph{functional} positional encoding, where the positional encodings are determined by some pre-defined functions (e.g., sinusoidal functions in~\cite{vaswani2017attention}). The other is the \emph{parametric} positional encodings, which are part of the model parameters and learned as in~\cite{devlin2019BERT}. ~\cite{shaw2018self} proposes a parametric \emph{relative} positional encoding, where the relative position information is incorporated in the self-attention layers of Transformer. Later, Transformer-XL~\cite{dai2019transformer} and XLNet~\cite{yang2019xlnet} propose using a sinusoid encoding matrix and two trainable bias terms to represent the relative positions. In this technical report, we employ a functional relative positional encoding scheme, which encodes the relative positions in self-attention by pre-defined functions without any trainable parameter. Our empirical study shows that it is an effective positional encoding scheme for the pre-trained language models, and it makes consistent gains in our experiments. Besides, we employed three techniques shown to be effective in the pre-training of the BERT model, which are \emph{Whole Word Masking}~\cite{cui2019pre}, \emph{Mixed Precision Training}~\cite{micikevicius2017mixed} and the \emph{LAMB Optimizer}~\cite{you2019reducing}, in training NEZHA.

The contribution of this technical report is that we systematically study the problem of pre-training language models on large-scale Chinese corpora, evaluate the models on several Chinese NLU tasks, and assess the effectiveness of training factors including positional encoding scheme, masking strategy, sources of training corpora, length of training sequences. We will release our NEZHA models as well as the source code to the community.

\section{Pre-training NEZHA Models}
\label{sec:encoding}

In this section, we present our NEZHA model in details. Section~\ref{sec:preliminary} presents the preliminaries of the BERT model and the positional encoding schemes. Section~\ref{sec:ourmodel} presents the functional relative positional encoding adopted in our model. Section~\ref{sec:wwm}, \ref{sec:mixed} and \ref{sec:lambopt} introduce the three techniques used in our pre-training, i.e., whole word masking, mixed precision training and the LAMB optimizer. 

\subsection{Preliminaries: BERT Model \& Positional Encoding}
\label{sec:preliminary}

BERT (Bidirectional Encoder Representations from Transformers) is a pre-trained language model, which is a stack of  Transformer encoders. Each Transformer encoder is a multi-head self-attention laryer followed by a position-wise feed-forward network. It uses residual connections around each sub-layer, followed by a layer normalization. We refer the reader to \cite{vaswani2017attention} for more details of the Transformer architecture. Each sample in the training data of BERT is a pair of sentences. In each sample, 12\% tokens are masked and 1.5\% tokens are randomly replaced by another token in the vocabulary. Besides, in the training set, each sample (containing sentences \emph{A} and \emph{B}) is constructed as follows. 50\% of the times, \emph{B} is actually the next sentence of \emph{A} and 50\% times \emph{B} is a random sentence from the corpus, which is not the next sentence of \emph{A}. In the pre-training phase, BERT has two tasks. One is the masked language modeling (MLM), which aims to predict the masked tokens from other tokens. The second pre-training task is the next sentence prediction (NSP). It predicts if the second sentence in each training sample is the next sentence of the first sentence or not. In some sense, BERT can be regarded as a denoising auto-encoder, since one of its training objectives is to recover the data with noises added. 

In Transformer, each attention head operates on a sequence of tokens $x=(x_1, x_2, \ldots, x_n)$, where $x_i\in \mathbb{R}^{d_{x}}$, and outputs a sequence $z=(z_1, z_2,\ldots, z_n)$ of the same length, where $z_i \in \mathbb{R}^{d_z}$. Each attention head has three parametric matrices $W^K, W^Q$ and $W^V \in \mathbb{R}^{d_x \times d_z}$ to be learned. The output $z_i$ is calculated as follows. 

\begin{equation}\label{eqn:z}
    z_i = \sum^n_{j=1}\alpha_{ij} (x_j W^{V} ).
\end{equation}

The attention score $\alpha_{ij}$ between the hidden states in position $i$ and position $j$ is computed by using a softmax function:

\begin{equation}\label{eqn:alpha}
    \alpha_{ij} = \frac{\exp{e_{ij}}}{\sum_{k}\exp{e_{ik}}},
\end{equation}

where $e_{ij}$ is the scaled dot product between the linear transformations of the input elements:

\begin{equation}\label{eqn:e}
    e_{ij} = \frac{(x_i W^Q)(x_j  W^K)^T}{\sqrt{d_z}}.
\end{equation}

Since the multi-head attention in Transformer (and BERT) is permutation invariant, and thus not sensitive to the word order. Therefore, \cite{vaswani2017attention} incorporates an absolute positional encoding for each position, which is an embedding vector and added to the token embedding directly. Later on, ~\cite{shaw2018self} proposes a parametric relative positional encoding for Transformer. In the relative positional encoding scheme, the computation of the attention scores involves a parametric embedding regarding the relative distance between the two positions. Specifically, it modifies the computation of the output $z_i$ in equation~\ref{eqn:z} and the $e_{ij}$ in equation~\ref{eqn:e} as follows: 

\begin{equation}
    z_i = \sum^n_{j=1}\alpha_{ij} (x_j W^{V} + a^V_{ij}),
\end{equation}

\begin{equation}
    e_{ij} = \frac{(x_i W^Q)(x_j  W^K + a^K_{ij})^T }{\sqrt{d_z}}.
\end{equation}

In the two equations above, $a^V_{ij}$, $a^K_{ij} \in \mathbb{R}^{d_z} $ are two vectors with the relative position between $i$ and $j$ encoded, and they are shared across all attention heads. Transformer-XL~\cite{dai2019transformer} and XLNet~\cite{yang2019xlnet} implement the relative positional encoding with a different formulation. We refer the reader to their paper for more details.

\subsection{Functional Relative Positional Encoding}
\label{sec:ourmodel}

In the current version of NEZHA, we employ functional relative positional encoding, where the computation of the outputs and attention scores involves sinusoidal functions of their relative position. This idea is inspired by the functional absolute positional encoding adopted in Transformer~\cite{vaswani2017attention}. Specifically, in our model, $a^V_{ij}$ and $a^K_{ij}$ are both derived from sinusoidal functions and fixed during the model training. In the remainder of this technical report, we denote $a_{ij}$ to present the formulation of $a^V_{ij}$ and $a^K_{ij}$ for clarity. Consider the dimension $2\cdot k$ and the dimension $2\cdot k +1$ of $a_{ij}$ respectively,

\begin{equation}
    a_{ij}[2k] = \sin((j-i)/(10000^{\frac{2\cdot k}{d_z}})),
 \end{equation}   
 \begin{equation}  
    a_{ij}[2k+1] = \cos((j-i)/(10000^{\frac{2\cdot k}{d_z}})).
\end{equation}

That is, each dimension of the positional encoding corresponds to a sinusoid, and the sinusoidal functions for different dimensions have different wavelengths. In the above equations, $d_z$ is equal to the hidden size per head of the NEZHA model (i.e., the hidden size divided by the number of heads). The wavelengths form a geometric progression from $2\pi$ to $10000 \cdot 2\pi$. We choose the fixed sinusoidal functions mainly
because it may allow the model to extrapolate to sequence lengths longer than the ones encountered during training. 

\subsection{Whole Word Masking}
\label{sec:wwm}

In the vanilla BERT, each token or Chinese character is masked randomly. In~\cite{cui2019pre}, whole word masking (WWM) strategy is found to be more effective than random masking for training BERT. In WWM, once a Chinese character is masked, the other characters belonging to the same Chinese word are all masked together. In implementing WWM for NEZHA, we used a tokenization tool Jieba\footnote{\url{https://github.com/fxsjy/jieba}} for the Chinese word segmentation (i.e., finding the boundaries of the Chinese words). In the WWM training data, each sample contains several masked Chinese words, and the total number of masked Chinese characters is roughly 12\% of its length and 1.5\% randomly replaced characters.



\subsection{Mixed Precision Training}
\label{sec:mixed}

In the pre-training of our NEZHA models, we adopt the technique of mixed precision training~\cite{micikevicius2017mixed}. The technique can speed up the training by 2-3 times and also reduce the space consumption of the model, as a result of which, a larger batch size could be utilized. 

Conventionally, the training of deep neural networks uses FP32 (i.e., single-precision float point format) to present all the variables (including the model parameters and gradients) involved in the training. Mixed precision training~\cite{micikevicius2017mixed} adopts \emph{mixed-precision} in the training. Specifically, it maintains a single-precision copy (called \emph{Master Weights}) of the weights in the model. In each training iteration, it rounds the Master Weights into FP16 (i.e., half-precision float point format) and performs the forward and backward pass with the weights, activations and gradients stored in FP16 format. Finally, it converts the gradients into FP32 format and updates the Master Weights by using the FP32 gradients. 

\subsection{LAMB Optimizer}
\label{sec:lambopt}

The LAMB optimizer~\cite{you2019reducing} is designed for the large batch-size synchronous distributed training of deep neuron networks. Training DNN with large mini-batches is an effective method to speed up the training. However, without careful tuning of the schedule of the learning rate, the performance could be largely harmed when the batch size is beyond a certain threshold. Instead of hand-tuning of the learning rate, the LAMB optimizer employs a general adaptation strategy and meanwhile provides insight into the convergence by theoretical analysis. The optimizer speeds up the training of BERT by using a very large batch size (up to more than 30k in \cite{you2019reducing}) without incurring a loss of the performance and even obtains the state-of-the-art performance in many tasks. Remarkably, the training time of BERT is significantly reduced from 3 days to 76 minutes. 

\section{Experiments}\label{sec:exp}

In this section, we report the experimental results on pre-training our NEZHA models for Chinese text and finetuning on Chinese NLU downstream tasks. It should be noted that the training techniques are not limited to Chinese and can be readily applied to other languages.

\subsection{Experimental Setting}

\paragraph{Datasets} We adopt three Chinese corpora for pre-training the NEZHA models:

\begin{itemize}
    \item Chinese Wikipedia~\footnote{\url{https://zh.wikipedia.org/wiki/}}. Chinese Wikipedia is a Chinese-language encyclopedia containing 1,067,552 articles. We downloaded the latest Chinese Wikipedia dump and cleaned the raw data with the tool named WikiExtractor\footnote{\url{https://github.com/attardi/wikiextractor}}. The cleaned corpus contains both simplified and traditional Chinese and has roughly 202M tokens.
    \item Baidu Baike~\footnote{\url{https://baike.baidu.com/}}. We crawled webpages from the Baidu Baike, which is a Chinese-language, collaborative, web-based encyclopedia owned and produced by the Chinese search engine Baidu. As of August 2018, Baidu Baike has more than 15.4 million articles. The cleaned corpus contains 4,734M tokens.
    \item Chinese News. We crawled Chinese News corpus from multiple news websites (e.g., Sina News). The cleaned corpus contains 5,600M tokens. 
\end{itemize}

For each corpus above, we prepared two versions of the pre-training data for NEZHA. The first version is processed the same as that in \cite{devlin2019BERT}, which contains 12\% masked Chinese characters and 1.5\% randomly replaced Chinese characters. We used tools provided by the BERT Github project~\footnote{\url{https://github.com/google-research/bert}} to convert the text data into the pre-training examples. The second version is based on the whole word masking (WWM) strategy. We created the WWM pre-training examples with the Chinese word segmenter Jieba for identifying the boundaries of Chinese words. In the WWM examples, each sample contains several masked Chinese words, and the total number of masked Chinese characters is roughly 12\% of its length and 1.5\% randomly replaced characters. Table~\ref{table:models} summarizes the statistics of the datasets for several pre-trained models.

\paragraph{Pre-training Details} We train the NEZHA models on 10 servers on Huawei Cloud~\footnote{\url{https://www.huaweicloud.com/product/modelarts.html}}, each of which has 8 NVIDIA Tesla V100 GPUs with 32GB memory. The distributed training algorithm is the \emph{Ring-AllReduce}\footnote{\url{https://github.com/baidu-research/baidu-allreduce}} and was employed with the framework named Horovod~\cite{sergeev2018horovod}. We trained each model from scratch and terminated the training when the training loss converged. For the NEZHA$_{\textsc{base}}$ models, we set the maximum learning rate to be $1.8e-4$ (with 1800 warm-up steps and linear decay). 
The batch size on each GPU is 180 and thus the total batch size is 180 * 8 * 10 = 14400. 
For the NEZHA$_{\textsc{large}}$ models, we set the maximum learning rate to be $1e-4$ (with 1800 warm-up steps and polynomial decay). 
The batch size on each GPU is 64, and thus the total batch size is 64 * 8 * 10 = 5120. 
In addition, we adopted the mixed-precision training using FP16~\cite{micikevicius2017mixed} in the pre-training phase. 

\begin{center}
\begin{table}[t]
\centering
\caption{Configurations of Chinese pre-trained language models}
\small{
\begin{tabular}{l p{2.7cm} cc p{1.5cm} p{2cm} p{2cm}}\hline
Model & Pre-Training Corpora & \#Tokens & Vocabulary size & Activation function & Hidden Size/\#Layers & \#Heads \\\hline
BERT$_{\textsc{base}}$ & \multirow{2}{2.7cm}{Wikipedia} & \multirow{2}{*}{202M} & \multirow{2}{2cm}{ 21,128}& \multirow{2}{2cm}{GeLU} & \multirow{2}{2cm}{768/12}& \multirow{2}{2cm}{12}\\
BERT$_{\textsc{base}}$-WWM & & & & &  & \\\hline
ERNIE-Baidu$_{\textsc{base}}$ 1.0 & Wikipedia+Baike+Tieba & 9,388 M & \multirow{3}{2cm}{18,000}& \multirow{3}{*}{ReLU}& 768/12& 12\\
ERNIE-Baidu$_{\textsc{base}}$ 2.0 & \multirow{2}{2.7cm}{Baike+News+Dialog} &  \multirow{2}{*}{14,988M} & & & 768/12& 12\\
ERNIE-Baidu$_{\textsc{large}}$ 2.0 & & & && 1024/24 & 16\\\hline
NEZHA$_{\textsc{base}}$ & \multirow{2}{2.7cm}{Wikipedia+Baike+News} & \multirow{2}{*}{10,536M} & \multirow{2}{2cm}{21,128} & \multirow{2}{1.5cm}{GeLU} & 768/12 & 12 \\
NEZHA$_{\textsc{large}}$ & &&&& 1024/24 & 16 \\ \hline
\end{tabular}
\label{table:models}}
\end{table}
\end{center}

\begin{center}
\begin{table}[t]
\centering
\caption{Pre-training Techniques Adopted in Chinese pre-trained language models (MLM: Masked Language Modeling, NSP: Next Sentence Prediction, WWM: Whole Word Masking, KM: Knowledge Masking, SR: Sentence Reordering, SD: Sentence Distance, DR: Discourse Relation, IR: IR Relevance, PAPE: Parametric Absolute Position Encoding, FRPE: Functional Relative Position Encoding)}
\fontsize{8.2}{9.2}\selectfont{
\begin{tabular}{l|ccc|c|p{1.2cm}|p{1.2cm}}\hline
\multirow{2}{*}{Model} & \multicolumn{3}{c|}{Pre-Training Tasks}  & \multirow{2}{*}{Training Precision} & \multirow{2}{*}{Optimizer} & \multirow{2}{1.2cm}{Position Encoding}  \\
 & Word-Aware & Sentence-Aware & Semantic-Aware &  &  &   \\\hline
BERT$_{\textsc{base}}$ & MLM & NSP &\multirow{2}{*}{-} & \multirow{2}{*}{Single Precision (FP32)} & ADAM & \multirow{2}{*}{PAPE} \\
BERT$_{\textsc{base}}$-WWM &MLM (WWM)& NSP& & & LAMB &\\\hline
ERNIE-Baidu$_{\textsc{base}}$ 1.0 & MLM (KM)& NSP & - & Single Precision (FP32) & ADAM& PAPE\\\hline
ERNIE-Baidu$_{\textsc{base}}$ 2.0&\multirow{2}{*}{MLM (KM)}& \multirow{2}{*}{SR \& SD}& \multirow{2}{*}{DR \& IR} &  \multirow{2}{*}{Mixed Precision} & \multirow{2}{*}{ADAM}& \multirow{2}{*}{PAPE} \\
ERNIE-Baidu$_{\textsc{large}}$ 2.0 &&& & & & \\\hline
NEZHA$_{\textsc{base}}$ &\multirow{2}{*}{MLM (WWM)}& \multirow{2}{*}{NSP}& \multirow{2}{*}{Span Prediction} & \multirow{2}{*}{Mixed Precision} & \multirow{2}{*}{LAMB} & \multirow{2}{*}{FRPE}  \\
NEZHA$_{\textsc{large}}$ & && && \\ \hline
\end{tabular}
\label{table:techniques}}
\end{table}
\end{center}

\subsection{Experimental Results}
\label{sec:exp_result}

In the experiment, we compared NEZHA models with the state-of-the-art Chinese pre-trained language models: Google's BERT~\cite{devlin2019BERT} for Chinese, BERT-WWM~\cite{cui2019pre} and ERNIE-Baidu~\cite{sun2019ernie,sun2019ernie2}. Their model configurations are shown in Table~\ref{table:models}. We also summarize pre-training techniques adopted in each Chinese pre-trained language models in Table~\ref{table:techniques}. Note that ERNIE-Baidu has three different versions, which are ERNIE-Baidu$_{\textsc{base}}$ 1.0, ERNIE-Baidu$_{\textsc{base}}$ 2.0 and ERNIE-Baidu$_{\textsc{large}}$ 2.0. ERNIE-Baidu$_{\textsc{base}}$ and ERNIE-Baidu$_{\textsc{large}}$2.0 introduced many different pre-training tasks and we refer the readers to their papers for the details of these tasks. As shown in the table, the unique technique in our models is the functional relative position encoding. We test the performances of the pre-trained models by finetuning on a variety of natural language understanding (NLU) tasks, which are listed as follows. The hyperparameters of finetuning each task are shown in Table~\ref{table:finetune}. 

\begin{itemize}
    \item \textbf{CMRC} (Chinese Machine Reading Comprehension 2018)~\cite{cui2018span}: A machine reading comprehension task that returns an answer span in a given passage for a given question.  
    \item \textbf{XNLI} (Cross-lingual Natural Language Inference)~\cite{conneau2018xnli}: The Chinese portion of XNLI, which is a version of MultiNLI where the dev and test sets have been translated (by humans) into 15 languages. XNLI is a natural language inference task. The goal of this task is to predict if the second sentence is a contradiction, entailment or neutral to the first sentence. 
    \item \textbf{LCQMC} (Large-scale Chinese Question Matching Corpus)~\cite{liu2018lcqmc}: A sentence pair matching task. Given a pair of sentences, the task is to determine if the two sentences are semantically equivalent or not. 
    \item \textbf{PD-NER} (People’s Daily Named Entity Recognition)~\footnote{\url{https://github.com/ProHiryu/bert-chinese-ner}}: A sequence labeling task that identifies the named entities from text. The corpus is from \emph{People's Daily}, a Chinese News Media.
    \item \textbf{ChnSenti} (Chinese Sentiment Classification)~\footnote{\url{https://github.com/pengming617/bert_classification}}: A binary classification task which predicts if the sentiment of a given sentence is positive or negative. 
\end{itemize}

\begin{center}
\begin{table}[h]
\centering
\caption{Hyperparameters used in finetuning downstream tasks. (SL: sequence length; LR stands: learning rate.)}
\begin{tabular}{ccccc|cccc}\hline
  Task Name &  Batch Size & SL & LR & Epochs & \#Train & \#Dev & \#Test & Domain \\
   &  (\textsc{base}/\textsc{large}) &  &  &  &   &  &  \\ \hline
  CMRC & 16/72 & 384 & 3e-5 & 2 & 10K & 3.2K & - & Wikipedia \\
  XNLI & 64/32 & 128 &3e-5 & 3 & 392K & 2.5K & 2.5K & General \\
  LCQMC & 64/32 & 128 &3e-5 & 5 & 240K & 8.8K & 12.5K & QA\\
  PD-NER & 64/16 & 256 &3e-5 & 5 & 51K & 4.6K & 68 & News\\
  ChnSenti & 64/16 & 256 & 3e-5 & 10 &9.6K & 1.2K & 1.2K  & General \\ 
   \hline
\end{tabular}
\label{table:finetune}
\end{table}
\end{center}

We show the comparison results of different pre-trained models on the aforementioned tasks in Table~\ref{table:result}. Among the groups of both \textsc{base} models and \textsc{large} models, either ERNIE-Baidu 2.0 or NEZHA achieves the best performances. Note that the part of the results are directly copied from the original papers~\cite{cui2019pre,sun2019ernie2}. Due to the possible differences in the experimental setting or finetuning methods, the comparison may not be entirely fair. We notice that there is consistent gaps between our implementation and the results reported in \cite{cui2019pre,sun2019ernie2} on the CMRC task. Once the ERNIE-Baidu 2.0 Chinese models are released, we will evaluate them under the same setting and update this report. 

\begin{center}
\begin{table}[t]
\centering
\caption{Results of pre-trained models on downstream Chinese NLU tasks.}
\begin{tabular}{p{3.5cm} cc cc cc cc cc }\hline
  Model &  \multicolumn{2}{c}{CMRC} & \multicolumn{2}{c}{XNLI} & \multicolumn{2}{c}{LCQMC} & \multicolumn{2}{c}{PD-NER} &\multicolumn{2}{c}{ChnSenti} \\
  & EM & F1 & Dev & Test & Dev & Test & Dev & Test & Dev & Test \\ \hline
  \textsc{BASE models}  \\
  BERT$_{\textsc{base}}$ & 64.06 & 85.01 & 78.75 & 77.27 & 89.04 & 87.61 & 96.53 & 98.58 &94.91 & 95.42\\
  BERT$_{\textsc{base}}$-WWM & 64.96 & 85.79 & 78.79 & 78.44 & 89.19 & 87.16 & 96.86 & 98.58 & 94.67 & 94.58\\
  BERT$_{\textsc{base}}$-WWM (in~\cite{cui2019pre}) & 66.30 & 85.60 & 79.00 & 78.20 & 89.40 & 87.00 & 95.30 & 65.10  & 95.10 & 95.40\\
  ERNIE-Baidu$_{\textsc{base}}$ 1.0 (in~\cite{sun2019ernie}) & 65.10 & 85.10 & 79.9 & 78.4 & 89.70 & 87.40 &- &- &95.20 & 95.40\\
  ERNIE-Baidu$_{\textsc{base}}$ 2.0 (in~\cite{sun2019ernie2}) & \textbf{69.10} & \textbf{88.60} & 81.20 & 79.70 & \textbf{90.90} & \textbf{87.90}  &- &- & \textbf{95.70} & 95.50 \\
  NEZHA$_{\textsc{base}}$ (ours) & 67.07 & 86.35 & \textbf{81.37} & 79.32 & 89.98 & 87.41 & 97.22 & 98.58 & 94.74 & 95.17\\
    NEZHA$_{\textsc{base}}$-WWM (ours) & 67.04 & 86.65 & 81.08 & \textbf{80.68} & 90.07 & 87.37 & \textbf{97.34} & \textbf{98.58} & 95.50 & \textbf{95.58}\\
      NEZHA$_{\textsc{base}}$-Span (ours) & 68.47 & 88.57 & 80.56 & 79.71 & 89.48 & 87.25 & 96.81 & 97.37 & 94.33 & 94.58 \\

  \hline
  \textsc{LARGE models}\\
  ERNIE-Baidu$_{\textsc{large}}$ 2.0 (in~\cite{sun2019ernie2}) & \textbf{71.50} & \textbf{89.90} & \textbf{82.60} & 81.00 & \textbf{90.90} & 87.90 & -&- & \textbf{96.10} & 95.80\\
  NEZHA$_{\textsc{large}}$ (ours) & 68.10 & 87.20 & 81.53 & 80.44 & 90.18 & 87.20 & \textbf{97.51} & \textbf{97.87} & 95.92 & 95.83 \\
  NEZHA$_{\textsc{large}}$-WWM (ours) &  67.32 & 86.62 & 82.21 & \textbf{81.17} & 90.87 & \textbf{87.94} & 97.26 & 97.63 & 95.75 & \textbf{96.00}\\
  \hline
\end{tabular}
\label{table:result}
\end{table}
\end{center}

\subsection{Ablation Study}
\label{sec:ablation}

In this section, we study the effectiveness of the data and different techniques for training NEZHA, which are listed as follows. 

\begin{itemize}
    \item \textbf{Positional Encoding: }the effectiveness of the functional relative positional encoding (FRPE) employed in our work compared with the parametric absolute positional encoding (PAPE) and parametric relative positional encoding (PRPE) adopted in the existing studies. 
    \item \textbf{Masking Strategy: }the effect of the whole word masking (WWM) on the performance of the pre-trained models.
    \item \textbf{Training Sequence Length: }the impact of the training with longer sequences.
    \item \textbf{Training Corpora: }the impact of the source of the training data.
\end{itemize}

With the above objectives, we compare the performances of several variants of NEZHA$_{\textsc{base}}$ model, as shown in Table~\ref{table:ablation}. The results demonstrate that the techniques mentioned above generally have positive contributions to the downstream tasks, where functional relative positional encoding shows a notable advantage compared with other positional encoding methods. For instance, we can see that when trained with a maximum of 128 tokens, the model using the absolute positional encodings  performs significantly worse than those using relative positional encodings on the CMRC task, where the input passages can be much longer.

\begin{center}
\begin{table}[h]
\centering
\caption{Ablation studies. (PAPE: parametric absolute positional encoding; PRPE: parametric relative positional encoding; FRPE: functional relative positional encoding; WWM: whole word masking; SL: sequence length.)}
{\footnotesize
\begin{tabular}{l |cc| cc| cc| cc| cc}\hline
  Model &  \multicolumn{2}{c|}{CMRC} & \multicolumn{2}{c|}{XNLI} & \multicolumn{2}{c|}{LCQMC} & \multicolumn{2}{c|}{PD-NER} &\multicolumn{2}{c}{ChnSenti}  \\ 
  & EM & F1 & Dev & Test & Dev & Test & Dev & Test & Dev & Test \\ \hline
  NEZHA$_{\textsc{base}}$ & & & & & & & & \\
			\hspace{3mm} \scriptsize{News, PAPE, SL:128} &  37.96 & 58.40&  78.79& 77.72 & 89.31 & 86.74 & 94.87 & 98.10 & 94.17& 95.67  \\
			\hspace{3mm} \scriptsize{News, PRPE, SL:128} &  65.26 & 86.17 &  79.18 & 77.98 &  89.21 & 86.92 & 96.93 & 98.12 & 94.67 & 95.08 \\
			\hspace{3mm} \scriptsize{News, FRPE, SL:128} &  65.95 & 86.46	& 79.96 & 78.32 & 89.40 & 87.23 & 96.69 & 98.10 &	95.58& 95.75  \\
			\hspace{3mm} \scriptsize{News, FRPE, SL:512} &  67.79 & 86.60 & 80.57 & 79.52 & 90.06 & 86.73 & 97.04 & 97.62 & 95.09 & 95.08 \\
			\hspace{3mm} \scriptsize{News+Wiki+Baike, FRPE, SL:128} &  66.95 & 86.41 & 81.25 & 79.06 & 89.83 & 87.13 & 97.21 & 97.41 & 95.25 & 94.42  \\
			\hspace{3mm} \scriptsize{News+Wiki+Baike, FRPE, WWM, SL:128} & 67.82 & 86.25 & 81.25 & 79.11 & 89.85 & 87.10 & 97.41 & 98.35 & 94.75 & 95.84 \\
			\hspace{3mm} \scriptsize{News+Wiki+Baike, FRPE, WWM, SL:512} & 66.45 & 86.16 & 80.96 & 79.86 & 89.64 & 86.18 & 96.79 & 98.10 & 95.08 & 95.42 \\
			\hspace{3mm} \scriptsize{News+Wiki+Baike, FRPE, Span, SL:128} & 68.47 & 88.57 & 80.56 & 79.71 & 89.48 & 87.25 & 96.81 & 97.37 & 94.33 & 94.58 \\
  \hline
\end{tabular}
\label{table:ablation}}
\end{table}
\end{center}

\section{Conclusion}
\label{sec:conl}

In the technical report, we have presented our practice on training the large scale pre-trained language models NEZHA on Chinese corpora. We have employed an effective functional relative positional encoding scheme, which leads to notable improvement over the other positional encodings. The pre-training of the NEZHA models also integrates several techniques, including whole word masking strategy, mixed precision training, and the LAMB optimizer. Experiments show that our models can achieve state-of-the-art performances on several Chinese natural language understanding tasks. In the future, we plan to continue the work on improving NEZHA on Chinese and other languages and extend the applications of NEZHA to more scenarios.

\bibliographystyle{unsrt}  
\bibliography{references}  

\begin{thebibliography}{10}

\bibitem{devlin2019BERT}
Jacob Devlin, Ming-Wei Chang, Kenton Lee, and Kristina Toutanova.
\newblock Bert: Pre-training of deep bidirectional transformers for language
  understanding.
\newblock In {\em Proceedings of the 2019 Conference of the North American
  Chapter of the Association for Computational Linguistics: Human Language
  Technologies, Volume 1 (Long and Short Papers)}, pages 4171--4186, 2019.

\bibitem{peters2018deep}
Matthew~E Peters, Mark Neumann, Mohit Iyyer, Matt Gardner, Christopher Clark,
  Kenton Lee, and Luke Zettlemoyer.
\newblock Deep contextualized word representations.
\newblock In {\em Proceedings of NAACL-HLT}, pages 2227--2237, 2018.

\bibitem{sun2019ernie}
Yu~Sun, Shuohuan Wang, Yukun Li, Shikun Feng, Xuyi Chen, Han Zhang, Xin Tian,
  Danxiang Zhu, Hao Tian, and Hua Wu.
\newblock Ernie: Enhanced representation through knowledge integration.
\newblock {\em arXiv preprint arXiv:1904.09223}, 2019.

\bibitem{sun2019ernie2}
Yu~Sun, Shuohuan Wang, Yukun Li, Shikun Feng, Hao Tian, Hua Wu, and Haifeng
  Wang.
\newblock Ernie 2.0: A continual pre-training framework for language
  understanding.
\newblock {\em arXiv preprint arXiv:1907.12412}, 2019.

\bibitem{zhang2019ernie}
Zhengyan Zhang, Xu~Han, Zhiyuan Liu, Xin Jiang, Maosong Sun, and Qun Liu.
\newblock Ernie: Enhanced language representation with informative entities.
\newblock {\em arXiv preprint arXiv:1905.07129}, 2019.

\bibitem{yang2019xlnet}
Zhilin Yang, Zihang Dai, Yiming Yang, Jaime Carbonell, Ruslan Salakhutdinov,
  and Quoc~V Le.
\newblock Xlnet: Generalized autoregressive pretraining for language
  understanding.
\newblock {\em arXiv preprint arXiv:1906.08237}, 2019.

\bibitem{Liu2019RoBERTaAR}
Yinhan Liu, Myle Ott, Naman Goyal, Jingfei Du, Mandar Joshi, Danqi Chen, Omer
  Levy, Mike Lewis, Luke Zettlemoyer, and Veselin Stoyanov.
\newblock Roberta: A robustly optimized bert pretraining approach.
\newblock 2019.

\bibitem{cui2019pre}
Yiming Cui, Wanxiang Che, Ting Liu, Bing Qin, Ziqing Yang, Shijin Wang, and
  Guoping Hu.
\newblock Pre-training with whole word masking for chinese bert.
\newblock {\em arXiv preprint arXiv:1906.08101}, 2019.

\bibitem{vaswani2017attention}
Ashish Vaswani, Noam Shazeer, Niki Parmar, Jakob Uszkoreit, Llion Jones,
  Aidan~N Gomez, {\L}ukasz Kaiser, and Illia Polosukhin.
\newblock Attention is all you need.
\newblock In {\em Advances in neural information processing systems}, pages
  5998--6008, 2017.

\bibitem{wu2016google}
Yonghui Wu, Mike Schuster, Zhifeng Chen, Quoc~V Le, Mohammad Norouzi, Wolfgang
  Macherey, Maxim Krikun, Yuan Cao, Qin Gao, Klaus Macherey, et~al.
\newblock Google's neural machine translation system: Bridging the gap between
  human and machine translation.
\newblock {\em arXiv preprint arXiv:1609.08144}, 2016.

\bibitem{shaw2018self}
Peter Shaw, Jakob Uszkoreit, and Ashish Vaswani.
\newblock Self-attention with relative position representations.
\newblock In {\em Proceedings of the 2018 Conference of the North American
  Chapter of the Association for Computational Linguistics: Human Language
  Technologies, Volume 2 (Short Papers)}, pages 464--468, 2018.

\bibitem{dai2019transformer}
Zihang Dai, Zhilin Yang, Yiming Yang, William~W Cohen, Jaime Carbonell, Quoc~V
  Le, and Ruslan Salakhutdinov.
\newblock Transformer-xl: Attentive language models beyond a fixed-length
  context.
\newblock {\em arXiv preprint arXiv:1901.02860}, 2019.

\bibitem{micikevicius2017mixed}
Paulius Micikevicius, Sharan Narang, Jonah Alben, Gregory Diamos, Erich Elsen,
  David Garcia, Boris Ginsburg, Michael Houston, Oleksii Kuchaiev, Ganesh
  Venkatesh, et~al.
\newblock Mixed precision training.
\newblock {\em arXiv preprint arXiv:1710.03740}, 2017.

\bibitem{you2019reducing}
Yang You, Jing Li, Jonathan Hseu, Xiaodan Song, James Demmel, and Cho-Jui
  Hsieh.
\newblock Reducing bert pre-training time from 3 days to 76 minutes.
\newblock {\em arXiv preprint arXiv:1904.00962}, 2019.

\bibitem{sergeev2018horovod}
Alexander Sergeev and Mike~Del Balso.
\newblock Horovod: fast and easy distributed deep learning in {TensorFlow}.
\newblock {\em arXiv preprint arXiv:1802.05799}, 2018.

\bibitem{cui2018span}
Yiming Cui, Ting Liu, Li~Xiao, Zhipeng Chen, Wentao Ma, Wanxiang Che, Shijin
  Wang, and Guoping Hu.
\newblock A span-extraction dataset for chinese machine reading comprehension.
\newblock {\em arXiv preprint arXiv:1810.07366}, 2018.

\bibitem{conneau2018xnli}
Alexis Conneau, Guillaume Lample, Ruty Rinott, Adina Williams, Samuel~R Bowman,
  Holger Schwenk, and Veselin Stoyanov.
\newblock Xnli: Evaluating cross-lingual sentence representations.
\newblock {\em arXiv preprint arXiv:1809.05053}, 2018.

\bibitem{liu2018lcqmc}
Xin Liu, Qingcai Chen, Chong Deng, Huajun Zeng, Jing Chen, Dongfang Li, and
  Buzhou Tang.
\newblock Lcqmc: A large-scale chinese question matching corpus.
\newblock In {\em Proceedings of the 27th International Conference on
  Computational Linguistics}, pages 1952--1962, 2018.

\end{thebibliography}




\end{document}